\documentclass{article}
\usepackage{spconf,amsmath,graphicx}
\usepackage[linesnumbered,ruled,lined]{algorithm2e}

\title{DTM: Deformable Template Matching}
\name{Hyungtae Lee, Heesung Kwon, Ryan M. Robinson, and William D. Nothwang}
\address{U.S. Army Research Laboratory, Adelphi, MD, USA}
\begin{document}
%
\maketitle
\begin{abstract}
A novel template matching algorithm that can incorporate the concept of deformable parts, is presented in this paper.  Unlike the deformable part model (DPM) employed in object recognition, the proposed template-matching approach called Deformable Template Matching (DTM) does not require a training step.  Instead, deformation is achieved by a set of predefined basic rules (e.g. the left sub-patch cannot pass across the right patch).  Experimental evaluation of this new method using the PASCAL VOC 07 dataset demonstrated substantial performance improvement over conventional template matching algorithms.  Additionally, to confirm the applicability of DTM, the concept is applied to the generation of a rotation-invariant SIFT descriptor.  Experimental evaluation employing deformable matching of SIFT features shows an increased number of matching features compared to a conventional SIFT matching.
\end{abstract}
\begin{keywords}
Template matching, deformable parts, SIFT
\end{keywords}
\section{Introduction}
\label{sec:intro}

Template matching refers to a set of techniques by which images are compared with a template image to find highly similar (matching) patches.  The similarity of patches is typically estimated in two major fashions, (i) pixel-to-pixel comparison and (ii) transformed comparison.  Sum of absolute difference (SAD)~\cite{DIBerneaTC72} and correlation~\cite{RBrunelli09,SKanekoPR03} are techniques belonging to the first category, and scale- and/or rotation-invariant matching~\cite{BSReddyTIP96,DGLoweIJCV04} (allowing affine transform between two image templates) belong to the second category.  However, less attention has been paid to template matching via deformations that are not explained by transformation.  Previous studies~\cite{PFFelzenszwalbCVPR07,JKimCVPR13} introduce elastic matching approaches by representing an image using a hierarchical tree structure.  However, the tree structure includes the whole image, preventing the use of small-sized template matching.

\begin{figure}[t]
\centering
\includegraphics[trim = 5mm 15mm 5mm 0mm,width=0.85\linewidth]{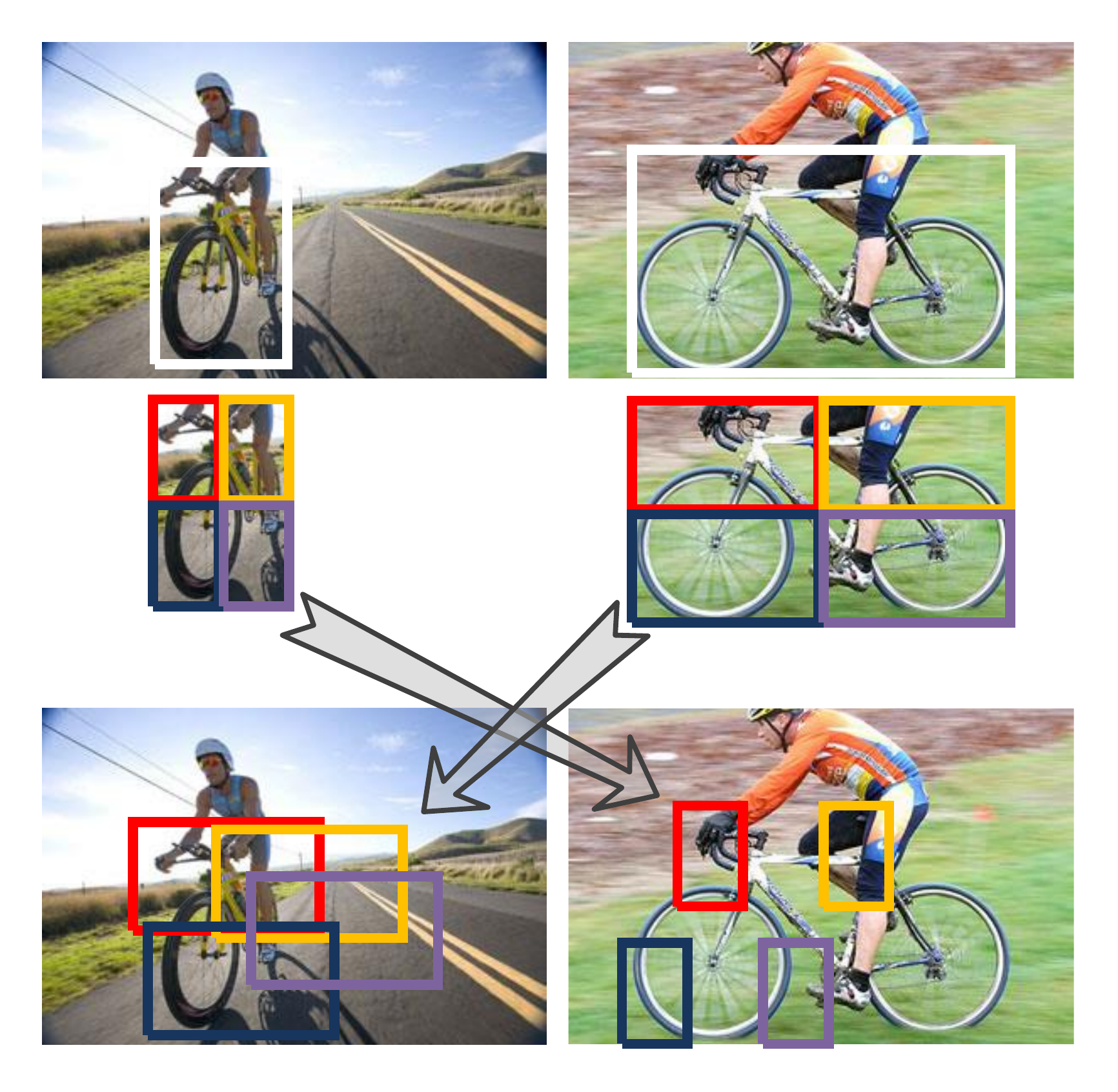}
\caption{{\bf Deformable Template Matching (DTM):} The first row shows two bicycle images taken from different viewpoints.  $2 \times 2$ decomposed sub-patches of bicycle images are in the second row.  In the third row, the sub-patches are matched to regions in the other image with the most similar shape.}
\label{fig:concept}
\end{figure}

Felzenswalb et al.~\cite{PFelzenszwalbPAMI10} introduced the concept of the deformable part model (DPM), which represents objects as a collection of basic parts arranged in a deformable configuration.  This representation is intuitive because it can describe, for instance, the different configurations of a person, whose joints and limbs do not move consistently, or the minor inconsistencies in the shape of similar objects, such as bicycles composed of the same parts (wheels, handles, etc.) held together by a similar but different frame.  DPM represents the state-of-the art in terms of object recognition task performance, however, traditional DPM requires a training set to learn deformation weights and cannot be directly applied to template matching.  Nevertheless, the concept of decomposing image patches into several sub-patches and allowing deformation among the sub-patches is applicable to template matching.  In fact, it extends template matching beyond ``exact'' matches, and expresses a deeper level of similarity than existing matching algorithms.  Highly similar patches can be of extremely high value in object recognition or when matching successive frames of an object in motion.  Figure~\ref{fig:concept} shows two images containing bicycles in different poses.  Note that placing sub-patches independently in the most similar locations of the other image leads us to conclude that the two images are from the same object.

This paper applies the concept of deformation to the task of matching two images.  Unlike other template matching approaches allowing for deformation, whose weights are learned over training images~\cite{AKJainPAMI96}, deformation of the location of sub-patches in this paper is restricted by a basic rule: once two sub-patches (e.g. on the left and right side of the image) are defined, the center location of the first (left) patch cannot pass across (be further right than) that of the second (right) patch.  The rule applies to left/right and top/bottom. Sub-patches can be placed at any position satisfying this rule.  The deformable template matching (DTM) process splits each image patch into several sub-templates and finds each patch a best match in the other image (so long as it satisfies the deformation rules through an iterative cost function).  The matching is also performed in the opposite direction by splitting the second image into sub-templates and applying them to the first image.  The overall matching score is a sum of the matching scores of both sets of sub-templates.

This paper first evaluates deformable template matching as a way of finding closely-matching image patches containing a specific object.  To confirm the strength of DTM in image processing applications, we also apply this concept to the generation of the well-known SIFT descriptor~\cite{DGLoweIJCV04}, and compare the performance against traditional means of generating the SIFT descriptor.  In Section~\ref{sec:exp}, the performance demonstrates that the proposed deformable template matching works effectively.


\section{Deformable Template Matching}
\label{sec:dtm}

Let $I_1$ and $I_2$ be two input images, where $I$ can be decomposed into $n \times m$ sub-patches, $I_p^{(i, j)},~i = 1, 2, \cdots , n,~j = 1, 2, \cdots , m$.  The sub-patch $I_p$ is matched at location $t = (x, y)$ in the other input image $I'$ with a matching cost $c(I_p, I'(t))$, where $I'(t)$ has the same width and height as the patch $I_p$.  Assume that two sub-patches (of image $I_1$), $I_{1,p}^{(i,j)}$ and $I_{1,p}^{(k,l)}$, are tentatively matched at location $t^{(i,j)}=(x^{(i,j)}, y^{(i,j)})$ and $t^{(k,l)}=(x^{(k,l)}, y^{(k,l)})$ in the coordinate frame of image $I_2$.  A deformation cost $d(t^{(i,j)}, t^{(k,l)})$ is defined as follows:

\begin{equation}
d(t^{(i,j)},t^{(k,l)}) = \left\{
\begin{array}{ll}
\infty & if~i>k,~y^{(i,j)}\leq y^{(k,l)}\\&~~~~~~or~ i<k,~y^{(i,j)}\geq y^{(k,l)}\\&~~~~~~or~ j>l,~x^{(i,j)}\leq x^{(k,l)}\\&~~~~~~or~ j<l,~x^{(i,j)}\geq x^{(k,l)}\\
0 & otherwise.
\end{array} \right.
\end{equation}

The deformation cost $d$ enforces hard constraints on the relative positioning of neighboring sub-patches, as mentioned in the previous section.  Suppose that two sub-patches, $I_p(1)$ and $I_p(2)$, originate as left and right (or up and down) in an image.  If these two patches are positioned in the other image in a manner that violates the original relative positioning, $d$ is given a value of infinity.

The total matching cost $c_{tot}$ between $I_1$ and $I_2$ is calculated as below:
\begin{equation}
c_{tot}(I_1, I_2) = c_{dtm}(I_1, I_2) + c_{dtm}(I_2, I_1),
\end{equation}
where $c_{dtm}$ is a score function of the deformable template matching that decomposes the first image into sub-patches and matches them to the second image.  $c_{dtm}$ consists of two terms: (i) the sum of the matching cost between each sub-patch and the second image and (ii) the sum of the deformation cost among sub-patches, minimized with respect to ${\bf t} = [t^{(1,1)},~\cdots,~t^{(n,m)}]$, as below:
\begin{eqnarray}
c_{dtm}(I_1, I_2) &=& \min_{t^{(1,1)},\cdots,t^{(n,m)}}\sum_{i=1}^n{\sum_{j=1}^m{c(I_{1,p}^{(i,j)},I_2(t^{(i,j)}))}}\nonumber\\
&&+\sum_{i,j,k,l:|i-k|\leq 1 \& |j-l|\leq 1}{d(t^{(i,j)},t^{(k,l)})}.
\end{eqnarray}\label{eq:cdtm}

\noindent In our implementation, we employ HOG features~\cite{NDalalCVPR05} to represent the image and compute the matching cost $c$ by using the sum of product of the features.

\begin{algorithm}
\caption{Proposed matching algorithm}
\label{alg:min_cdtm}
\SetKwFunction{Decompose}{Decompose}
\SetKwFunction{Initialization}{Initialization}
\SetKwFunction{MinCdtm}{MinCdtm}

\KwIn{$I_1,~I_2$}
\KwOut{${\bf t},~c$}
\BlankLine
$I_{1,p}~\gets$ \Decompose($I_1,~n,~m$)\;
${\bf t}~\gets$ \Initialization($I_2,~n,~m$)\;
s $\gets$ 1\;
\While{1}{
${\bf t}_{old} \gets {\bf t}$\;
c $\gets$ 0\;
\For{i=1 \emph{\KwTo} n}{
\For{j=1 \emph{\KwTo} m}{
$[t^{(i,j)},c^{(i,j)}]~\gets~~~~~~~~~~~~~~~~~~~~~~~~~~~$   $~~~~~$\MinCdtm($t^{(i,j)},s,{\bf t},I_{1,p}^{(i,j)},I_2$)\;
$c~\gets~c+c^{(i,j)}$\;
}
}
\BlankLine
\If{${\bf t}_{old}=={\bf t}$}{
break\;
}
\BlankLine
s $\gets$ s+1\;
}
\end{algorithm}

Algorithm~\ref{alg:min_cdtm} presents the implementation of $c_{dtm}$ computation.  The location of sub-patches, ${\bf t}$, are obtained by minimizing the cost function given by Equation~\ref{eq:cdtm}.  The function {\tt Decompose}($I$, $n$, $m$) decomposes an image $I$ into $n \times m$ sub-patches.  {\tt Initalization}($I_2$, $m$, $n$) is a function setting the initial position of the sub-patches in the other image.  The initial patch center location $t^{(i,j)}=(x^{(i,j)},y^{(i,j)})$ is calculated as $(\lfloor (i-1)\times I_2\_width/n \rfloor,~\lfloor (j-1)\times I_2\_height/m\rfloor)$, $i=1,\cdots,n,~j=1,\cdots,m$.  The function {\tt MinCdtm} searches $t^{(i,j)}$ minimizing $c_{dtm}$ in a range $[x^{(i,j)}-s~x^{(i,j)}+s,~y^{(i,j)}-s~y^{(i,j)}+s]$, where $s$ is a search area.  The minimization is achieved by increasing the search area $s$ in every iteration.  If ${\bf t}$ is not changed, the process terminates.


\section{Deformable SIFT Matching}
\label{sec:def_sift_descr}


To emphasize the applicability of deformable template matching, we apply the concept in feature matching as well as template matching.  While template matching is for measuring how similar given two templates are, feature matching is used to find a transformation between two images by assuming one image is transformed from other image in some fashion.

DTM can be used in matching any type of feature, thus we modify the well-known SIFT descriptor by adding deformability to the existing rotation-invariant methods.  The SIFT descriptor consists of 4x4 cells, each of which collects the magnitude of 8 gradients. The magnitude of gradients is calculated after rotating the neighboring region around each keypoint so that the dominant gradient of all keypoints faces in the same direction.  The similarity of two SIFT keypoints are calculated as sum of product of their 128-dimensioned descriptor.  The proposed matching reconfigures $4 \times 4$ cells of the descriptor to $2 \times 2$ sub-patches and applies deformable template matching to compute the similarity between two different SIFT keypoints.  Gaussian smoothing used in the traditional SIFT descriptor is not applied due to the variable locations of sub-patches, which allows direct matching of two SIFT descriptors.

\section{Experiments}
\label{sec:exp}

\subsection{Comparison with Other Matching Techniques}
\noindent {\bf Dataset and setting:} The PASCAL VOC 07 image dataset~\cite{pascal-voc-2007} was used to evaluate the proposed deformable template matching against several existing techniques. However, we did not follow the protocol of PASCAL VOC 07 because it is intended for the task of object recognition, rather than template matching.  A single object patch in one of the images is randomly selected from all of the object patches in the dataset, cropped by annotated bounding boxes.  Then, 100 positive and negative patches are randomly selected from the same object category and different object categories, respectively.  This procedure iterates 100 times and in each iteration, area under ROC curves (AUC) is calculated.


\noindent {\bf Baseline:} First, SAD is used as the baseline.  Since two templates can have different sizes, matching via SAD is performed in two ways: (i) transform one image so that it is equally sized with the other image and compute SAD (SAD1), and (ii) scan one image over other image to search for the maximum matching score (SAD2).  The matching calculations are also performed in the opposite direction by switching the first and second templates with each other, and the sum of the matching scores is computed in both directions to obtain the final score.  As a second baseline, the images are converted via HOG~\cite{NDalalCVPR05} features and the matching score is computed by sum of product of HOG features of two templates. HOG feature-based matching is performed bi-directionally (HOG1 and HOG2), as described with SAD.  SIFT features are not used in this baseline comparison because without Gaussian smoothing SIFT shares the same principle of exploiting gradient magnitudes and orientations as HOG.


\begin{figure}[t]
\centering
\includegraphics[trim = 0mm 0mm 0mm 15mm,width=0.9\linewidth]{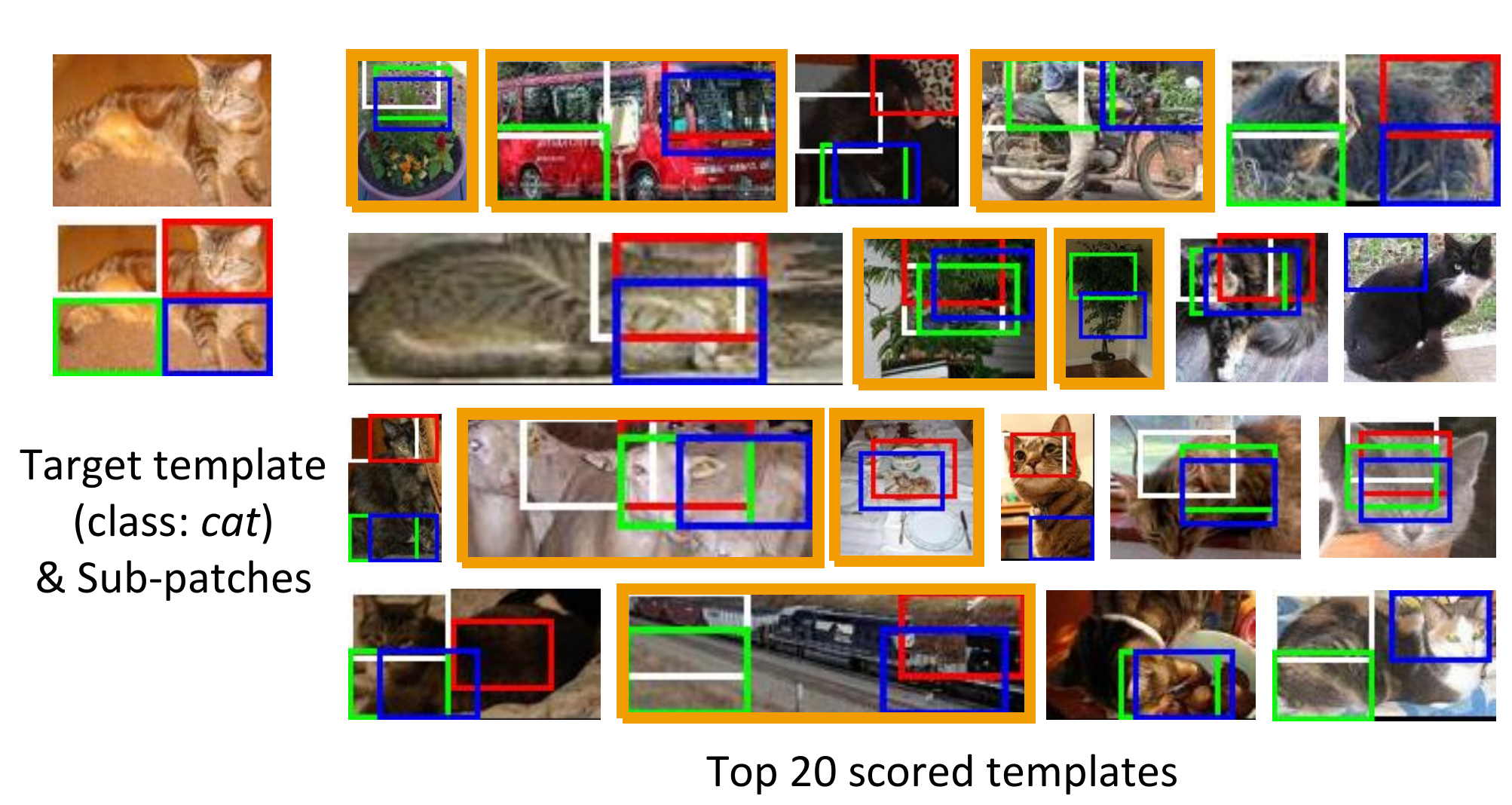}\\
\includegraphics[trim = 8mm 15mm 8mm 5mm,width=0.9\linewidth]{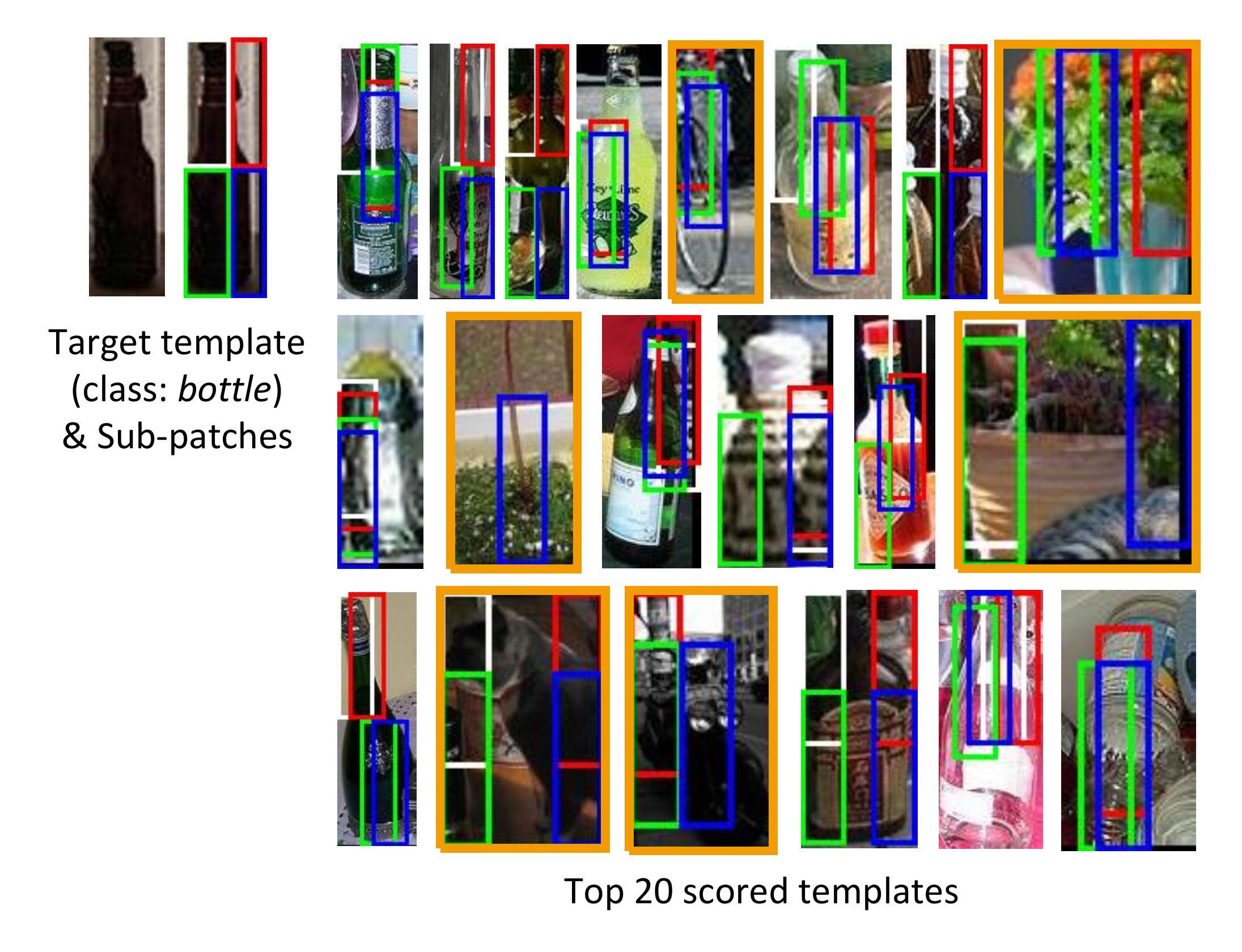}
\caption{{\bf Qualitative analysis:} The left-most figures are target templates (${\it cat}$ \& ${\it bottle}$) and their 2$\times$2 sub-patches are shown below (or beside).  Among 200 positive and negative templates, the top 20 matching scored images are shown on the right. The images boxed in orange are negative templates.}
\label{fig:qual_anal}
\end{figure}

\begin{table}[t]
\centering
\caption{Performance of DTM with various number of sub-patches. (Mean and standard deviation of AUC)}
\begin{tabular}{cccc}
\hline
\# of sub-patches & $2 \times 2$ & $3 \times 3$ & $4 \times 4$ \\\hline
mean & .6358 & .6264 & .6079 \\
std & .1258 & .1486 & .1485 \\
\hline
\end{tabular}
\label{tab:num_of_subpatch}
\end{table}

\begin{table}[t]
\centering
\caption{Comparison between DTM and baselines. (Mean and standard deviation of AUC)}
\begin{tabular}{cccccc}
\hline
method & SAD1 & SAD2 & HOG1 & HOG2 & DTM \\\hline
mean & .5430 & .5419 & .6178 & .6150 & .6358 \\
std & .1245 & .1213 & .1355 & .1311 & .1258 \\
\hline
\end{tabular}
\label{tab:auc}
\end{table}


Figure~\ref{fig:qual_anal} shows two randomly selected templates from the object categories (${\it cat}$ and ${\it bottle}$), their $2 \times 2$ sub-patches, and the top 20 best matching templates.  Among the top 20 templates, negative templates are also included (eight for ${\it cat}$, six for ${\it bottle}$) when using DTM.    Note that using DTM, the sub-patch containing the cat head (red box) is accurately positioned over the other cat heads in the corresponding templates.  Table~\ref{tab:num_of_subpatch} evaluates matching performance of DTM as the number of sub-patches is varied.  DTM with $2 \times 2$ sub-patches works best among others due to the already low-resolution of the templates.  Table~\ref{tab:auc} compares the four baselines and DTM.  Based on these characteristics, DTM outperforms all the baselines.

\subsection{Evaluation for Deformable SIFT Matching}
\label{ssec:eval_deform_sift_descr}

\noindent {\bf Data and setting:} We use the ``Lena'' image to evaluate the proposed deformable feature matching based on the SIFT descriptor.  The image is transformed with respect to arbitrarily-selected rotation and scale along the $x$ and $y$ axes.  RANSAC~\cite{MAFischlerCACM81} is employed to search for inliers and homography between the original Lena image and the transformed image.  This process is performed 100 times.  If homography properly reproject the transformed image to the original, more correct matchings (inliers) implies better matching.

\begin{figure}[t]
\centering
\includegraphics[trim = 5mm 25mm 5mm 25mm,width=0.9\linewidth]{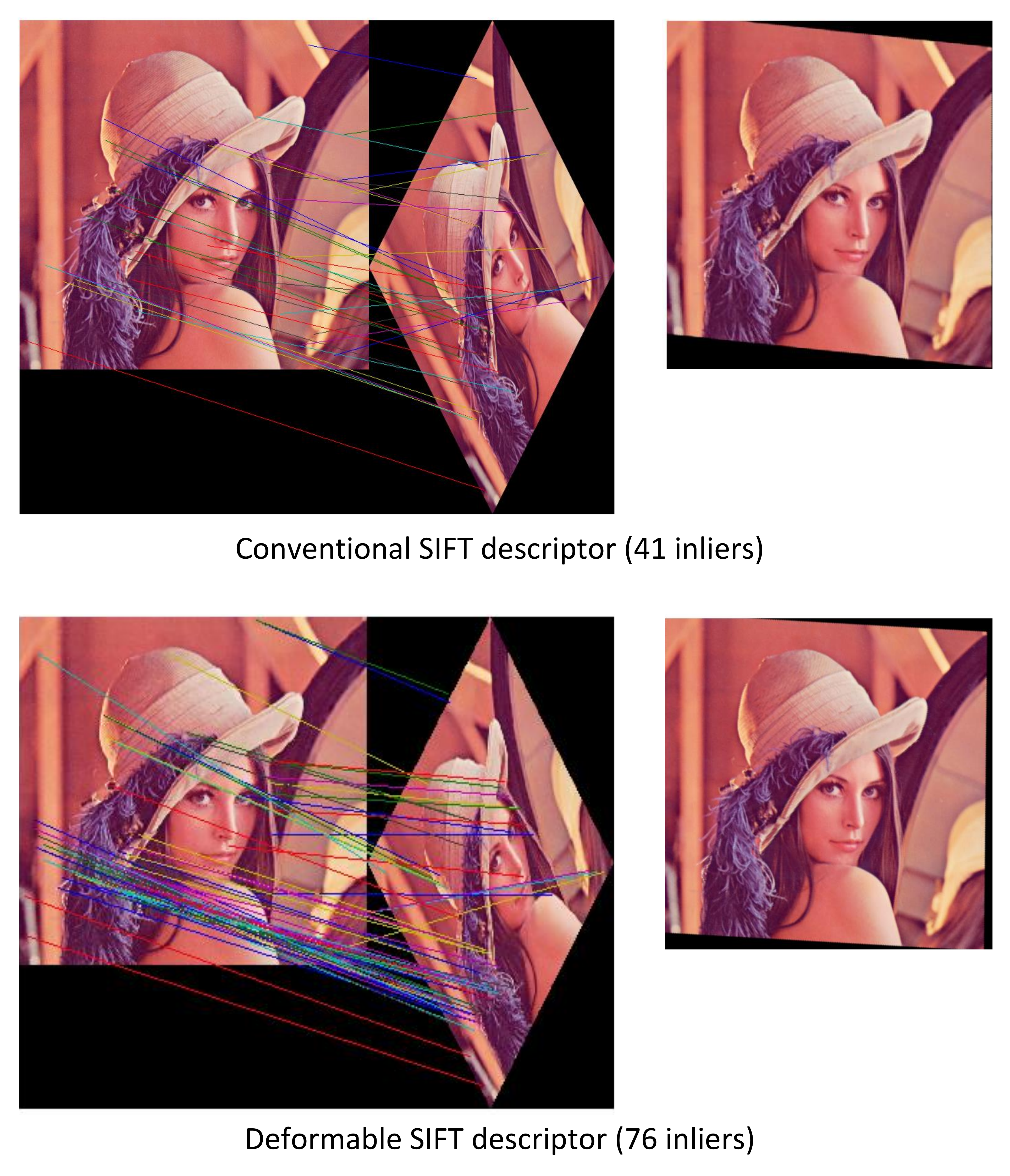}
\caption{{\bf Deformable SIFT matching vs conventional SIFT matching:} First column shows matching inliers and second column shows projected $2^{nd}$ image by computed homography through RANSAC.  $2^{nd}$ image is obtained by rotating the Lena image by $\pi/2$ anti-clockwise and resize 0.5 w.r.t. x-axis.}
\label{fig:def_sift_descr}
\end{figure}

Figure~\ref{fig:def_sift_descr} shows the comparison between the deformable SIFT matching ($2^{nd}$ row) and conventional SIFT matching ($1^{st}$ row).  Based on the projected image (right side), we can see that both matching techniques find the proper homography.  However, deformable SIFT matching finds more inliers than conventional SIFT matching.  Table~\ref{tab:num_inlier} summarizes mean and standard deviation values for the multiple trials, demonstrating that the proposed deformable SIFT matching finds more properly-matching SIFT features than the conventional SIFT descriptor.  This implies that deformable SIFT matching is a more reliable method for finding relationships between two images.



\begin{table}[t]
\centering
\caption{$\#$ of inlier matching}
\begin{tabular}{ccc}
\hline
& conv. SIFT & deform. SIFT \\\hline
mean & 153.4 & 223.9 \\
std & 21.54 & 24.54 \\
\hline
\end{tabular}
\label{tab:num_inlier}
\end{table}

\section{Conclusion}
\label{sec:concl}

This work presented a new template matching technique called deformable template matching (DTM).  Unlike other template matching methods, DTM is able to account for image or object deformations that are not caused by transformation, enabling a greater flexibility to find similar objects or features.  Although DTM is conceptually similar to the deformable part model (DPM) employed in object recognition, no training is required to perform template matching. Instead, a rule defining the relative locations of deformed sub-patches allows for deformable matching.

DTM was experimentally tested and compared to various baseline methods using images from the PASCAL VOC 07 dataset.  A quantitative analysis of receiver operating characteristic (ROC) indicated that DTM performed better on average than baseline versions of sum of absolute difference (SAD) and histogram of oriented gradients (HOG) methods in matching image patches featuring objects of the same category.  Additionally, the deformable SIFT matching was directly compared to a conventional SIFT matching. The deformable SIFT matching produced more inlier matches, suggesting better re-projection is possible with the proposed method.

\section{Acknowledgement}

 This project was supported by the U.S. Army Research Laboratory under a Director's Strategic Research Initiative entitled "Heterogeneous Systems for Information Variable Environments (HIVE)" from FY14-FY16. The views and conclusions contained in this document are those of the authors and should not be interpreted as representing the official policies, either expressed or implied, of the Army Research Laboratory or U.S. Government. The U.S. Government is authorized to reproduce and distribute reprints for Government purposes notwithstanding any copyright notation herein.


\bibliographystyle{IEEEbib}
\bibliography{strings,refs}

\end{document}